\documentclass{article}
\usepackage{arxiv}

\usepackage{comment}
\usepackage{booktabs} 
\usepackage{multirow}
\usepackage{tikz}
\usepackage{graphicx}
\usepackage{makecell}

\usepackage{subcaption}
\usepackage{amsmath}
\usepackage{algorithm}
\usepackage[noend]{algpseudocode}
\usepackage{xfrac}
\usepackage[labelfont=bf]{caption}
\usepackage[capitalize,noabbrev]{cleveref}

\graphicspath{{figs/}} 
\newcommand{\sys}{AdaTest}

\title{\sys{}: Reinforcement Learning and Adaptive Sampling for On-chip Hardware Trojan Detection}

\date{}

\author{Huili Chen \\
	UC San Diego\\
	La Jolla, CA \\
	\texttt{huc044@ucsd.edu} \\
	\And
	Xinqiao Zhang \\
	UC San Diego  \& San Digeo State University\\
	La Jolla, CA \\
	\texttt{x5zhang@ucsd.edu}
	\And
	Ke Huang \\
	San Diego State University\\
	San Diego, CA \\
	\texttt{khuang@sdsu.edu}	
	\And
	Farinaz Koushanfar \\
	UC San Diego\\
	La Jolla, CA \\
	\texttt{farinaz@ucsd.edu}
}

\begin{document}
\maketitle



\begin{abstract}
This paper proposes \sys{}, a novel \textit{adaptive test pattern generation} framework for efficient and reliable Hardware Trojan (HT) detection. HT is a backdoor attack that tampers with the design of victim integrated circuits (ICs).
\sys{} improves the existing HT detection techniques in terms of scalability and accuracy of detecting smaller Trojans in the presence of noise and variations.
To achieve high trigger coverage, \sys{} leverages Reinforcement Learning (RL) to produce a {diverse} set of test inputs. Particularly, we \textit{progressively} generate test vectors with high `reward' values in an \textit{iterative} manner. In each iteration, the test set is evaluated and adaptively expanded as needed.  
Furthermore, \sys{} integrates \textit{adaptive sampling} to prioritize test samples that provide more information for HT detection, thus reducing the number of samples while improving the samples' quality for faster exploration. 
We develop \sys{} with a \textit{Software/Hardware co-design} principle and provide an optimized on-chip architecture solution. \sys{}'s architecture minimizes the hardware overhead in two ways: (i) Deploying circuit emulation on programmable hardware to accelerate reward evaluation of the test input; (ii) Pipelining each computation stage in \sys{} by automatically constructing auxiliary circuit for test input generation, reward evaluation, and adaptive sampling. 
We evaluate \sys{}'s performance on various HT benchmarks and compare it with two prior works that use logic testing for HT detection. 
{Experimental results show that \sys{} engenders up to two orders of test generation speedup and {two orders of test set size reduction} compared to the prior works while achieving the same level or higher Trojan detection rate. } 
\end{abstract}



\maketitle

\section{Introduction}   \label{sec:intro}

Integrated circuits (ICs) are indispensable components for a diverse set of real-world applications including healthcare systems, smart home devices, industrial equipment, and machine learning accelerators~\cite{chen2016eyeriss, chen2009wireless}. 
The vulnerability of digital circuits may result in severe outcomes due to their deployment in security-critical tasks.
The design and manufacturing process of contemporary ICs are typically outsourced to (untrusted) third parties. 
Such a supply chain structure results in hardware security concerns, such as sensitive information leakage, performance degradation, and copyright infringement~\cite{tehranipoor2011introduction,colombier2014survey}.
Malicious hardware modifications, a.k.a., \textit{Hardware Trojan (HT)} attack~\cite{tehranipoor2010survey,bhunia2014hardware}
may occur at each stage of the IC supply chain. 

There are two main components in a HT attack: Trojan trigger and payload. 
The HT \textit{trigger} is a control signal that determines when the malicious activity of the HT shall be activated. 
The Trojan \textit{payload} is the actual effect of circuit malfunctioning which depends on the purpose of the adversary, e.g., stealing private information or producing incorrect outputs~\cite{tehranipoor2010survey}.
The attacker intends to design a stealthy HT that remains dormant during functional testing and evades possible detection techniques. 
As such, the HT trigger is typically derived from the rather rare activation conditions that are easier to hide for the intruder.

To alleviate the concerns about malicious hardware modifications, a line of research has focused on developing effective HT detection methods. 
Existing HT detection techniques can be categorized into two classes based on the underlying mechanisms: \textit{(i) Side-Channel Analysis} (SCA), and, \textit{(ii) Logic Testing}. 
SCA-based HT detection explores the fact that the presence of the HT on the victim circuit will change its \textit{physical parameters} (e.g., time, power, and electromagnetic radiation), thus can be revealed by side-channel information~\cite{liu2014hardware,lin2009trojan}. 
Such a mechanism determines that SCA-based approaches can detect \textit{non-functional} HTs, while they may have high false alarm rates when detecting small HTs due to the operational and physical silicon variation, as well as measurement noise. 
Logic testing-based techniques intend to activate the stealthy Trojan trigger by generating diverse test patterns~\cite{chakraborty2009mero,nourian2018hardware,saha2015improved}. 
The main challenge of logic testing-based HT detection is to increase the \textit{trigger coverage} with a small number of test patterns. 

\begin{figure}[ht!]
    \centering
    \includegraphics[width=0.9\textwidth]{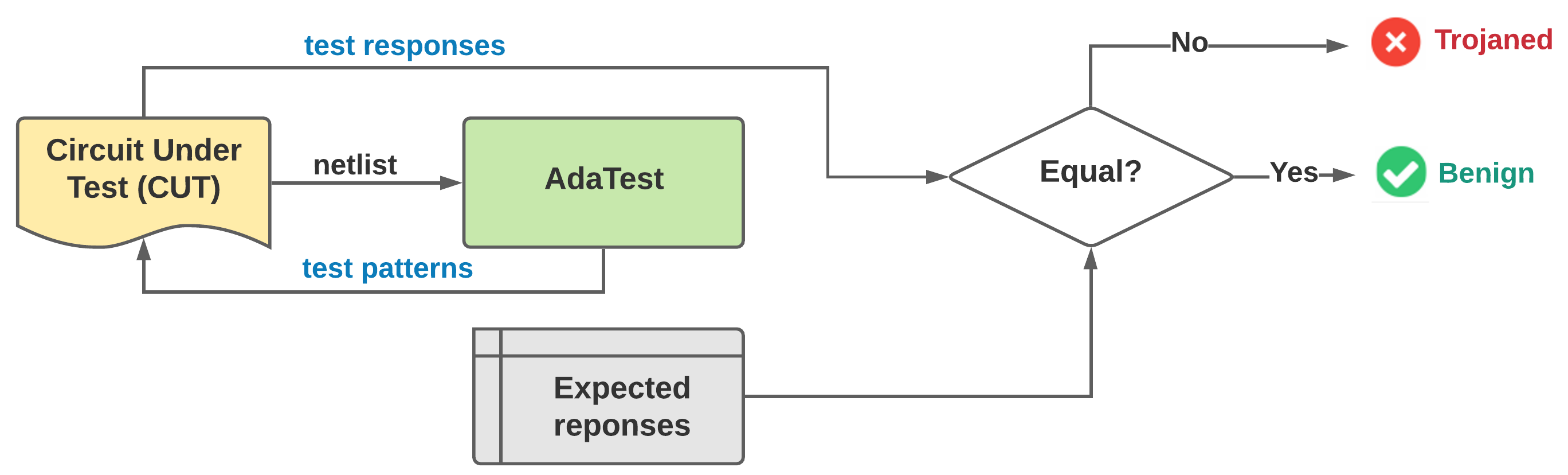} 
    \vspace{-0.2em}
    \caption{High-level usage of \sys{} for hardware-assisted security assurance against Trojan attacks.}
    \label{fig:sys_demo}
    \vspace{-0.7em}
\end{figure}

In this paper, we aim to simultaneously address three challenges of logic testing-based HT detection: effectiveness, efficiency, and scalability. 
To this end, we propose~\sys{}, the first automated \textbf{adaptive, reinforcement learning-based test pattern generation (TPG)} framework for HT detection with \textit{hardware accelerator design}.
Figure~\ref{fig:sys_demo} demonstrates the high-level usage of \sys{} to inspect if any hardware Trojans are inserted in the CUT. 
\sys{} takes the netlist of the circuit under test (CUT) and user-defined parameters as its inputs. A set of test vectors with high reward values are returned as the output of \sys{}.

\sys{} framework consists of two main phases: \textbf{(i) Circuit profiling}. Given the circuit netlist, we first characterize each node in the CUT from two perspectives: the \textit{transition probability}, and the \textit{SCOAP testability} measures. These two properties are used to identify rare nodes and quantify the fitness of each node, respectively. 
\textbf{(ii) Adaptive test pattern generation.} \sys{} proposes an innovative reward function for test vectors using the following information: the number of times that each rare node is triggered, the SCOAP testability measure of the rare nodes, and the graph-level distance of the circuit (represented as directed acyclic graph) when applying this test input and the historical ones. In each iteration, \sys{} gradually expands the test set by generating candidate test inputs and selecting the ones that have high reward values. \sys{} provisions a flexible \textit{trade-off} between trigger coverage and test generation time. 
To enable hardware-assisted solution, we further design an optimized architecture for \sys{}'s implementation to reduce the hardware overhead. 
More specifically, \sys{} architecture pipelines the computation in online TPG and deploy circuit emulation to accelerate reward evaluation.   
\vspace{-1em}

\sys{} opens a new axis for the growing research in hardware security by exploring
the idea of reinforcement learning (RL) and adaptive test pattern generation.
The adaptive nature of \sys{} ensures that the quality (measured by our reward function)
of our dynamic test set always improves over iterations as new test inputs are added to the test set. 
Furthermore, \sys{} is \textit{generic} and can be easily extended for other hardware security problems, such as {logic verification, efficient ATPG, functional testing, and built-in self-test.} For example, the concept of RL and adaptive test pattern generation presented in \sys{} can be used in an efficient ATPG application where the RL reward function is designed to reflect the goal of the ATPG (such as fault coverage of considered fault models).      


\noindent \textbf{Organization. }
Section~\ref{sec:prelim} introduces preliminary knowledge and related works on Hardware Trojan and its detection, as well as reinforcement learning. Section~\ref{sec:overview} discusses the challenges of HT detection and the overall workflow of \sys{} framework. Section~\ref{sec:alg} presents our test pattern generation algorithm that combines RL and adaptive sampling for fast exploitation.    
Section~\ref{sec:hardware} demonstrates our domain-specific architecture design of \sys{}. Section~\ref{sec:eval} provides a comprehensive performance evaluation of \sys{} on various circuit benchmarks and comparison with prior works on logic testing-based HT detection. 
Section~\ref{sec:conclus} concludes the paper. 
\section{Preliminaries and Backgrounds} \label{sec:prelim}

\subsection{Hardware Trojan Attacks} \label{sec:HT_attacks}
\begin{figure}[ht!]
    \centering
    \includegraphics[width=0.4\textwidth]{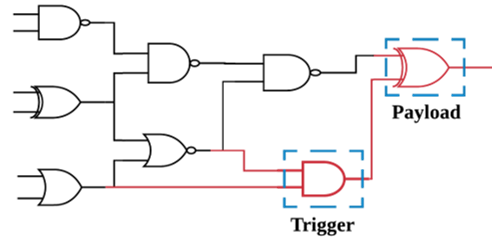} 
    \vspace{-0.5em}
    \caption{Demonstration of the Hardware Trojan attack.}
    \label{fig:HT_demo}
    \vspace{-0.5em}
\end{figure}

Security of third-party SoCs has raised an increasing amount of concerns due to the contemporary outsourcing-based supply chain. 
Hardware Trojans are malicious circuit modifications inserted in the circuit to perform the pre-defined adversarial task (`payload') e.g., circuit malfunction or private information leakage, when its control signal (`trigger') is activated. 
Figure~\ref{fig:HT_demo} shows an example HT design where a logic-AND gate and an XOR-gate are used as the trigger and payload, respectively. 
The payload flips the output signal when the trigger is activated, thus disturbing the desired behavior of the original circuit. 

\vspace{-0.8em}

The collaborative nature of the supply chain also determines that HTs may be inserted by different parties at different stages of the IC lifecycle. For instance, the untrusted IP provider, the circuit designer, or the manufacturing party might insert HTs in the circuit. 
Hardware Trojans shall remain \textit{dormant} in most cases to evade functional testing and HT detection, while it should be activated by the trigger to execute the attack. 
For this purpose, stealthy HTs are designed with two main considerations: (i) Rare conditions are used to construct the trigger signal; (ii) The HT is placed in a non-critical path to minimize its impact on side channels (delay, power, electromagnetic emission, etc.)

\vspace{-0.5em}
\subsection{Hardware Trojan Detection} \label{sec:HT_detect}
\vspace{-0.4em}
Previous HT detection techniques can be categorized into two broad types: destructive and non-destructive methods. 
Destructive detection schemes perform de-packaging and de-layering on the manufactured IC to reverse engineer its design layout, thus is prohibitively expensive~\cite{el2015integrated}.
Non-destructive HT detection includes two types: run-time monitoring and test-time detection.
Run-time approaches monitor the IC throughout its entire operational lifecycle with the goal of detecting Trojans that pass other detection methods, providing the 'last-line of defense'. 
There are two classes of test-time HT detection techniques. We detail each type as follows: 

\textbf{(i) Side-channel Analysis.} 
SCA-based Trojan detection methods explore the influence of the inserted HT on a particular measurable physical property, such as the supply current, power consumption, or path delay. 
These physical traces can be considered as the \textit{`fingerprint'} of the circuit and allow the defender to detect both \textit{parametric} and \textit{functional} Trojans~\cite{liu2015concurrent,liu2014hardware}.
{Parametric Trojans modify the wires and/or logic in the original circuit while functional Trojans add/delete transistors or gates in the original chip~\cite{wang2008hardware, karri2012trojan,moein2015attribute}.}
However, SCA-based HT detection has two limitations: (i) It cannot detect a small HT that causes a negligible impact on the physical side-channel; (ii) The extracted circuit fingerprint is susceptible to manufacturing variation and measurement noise, thus it might incur high false alarm rates.

\textbf{(ii) Logic Testing.} 
Compared to the side-channel based approaches, logic testing methods can only detect \textit{functional} Trojans.
However, they yield reliable results under process variation and measurement noise. 
The main challenge of developing a practical and effective logic testing technique for HT detection is the inordinately large space of possible Trojan designs that the adversary can explore. 
Since the HT trigger is derived from a very rare condition that is unknown to the defender, attempting to stimulate the stealthy Trojan with a limited number of test inputs is difficult. 
Existing logic testing methods generate test patterns using simple heuristics, thus cannot ensure high trigger coverage on complex circuits.
Also, such heuristic-driven test generation approaches are inefficient (long test generation time) and unscalable to large benchmarks~\cite{chakraborty2009mero,bhunia2014hardware,tehranipoor2010survey}.

Besides SCA and logic testing, other HT detection techniques have also been explored. For instance, FANCI~\cite{waksman2013fanci} presents a Boolean functional analysis method to identify suspicious wires that are nearly unused in the circuit. For this purpose, FANCI introduces a concept called `control value' to characterize the influence of a specific wire on other wires. The wires with small control values are flagged as suspicious. 
{However, the wire-wise control value computation in FANCI is unscalable on large circuits.}
VeriTrust~\cite{zhang2015veritrust} suggests a verification method to detect HT trigger inputs by examining the verification corners. Therefore, VeriTrust is agnostic to the HT implementation styles.

Prior works on logic testing have explored various heuristics to improve trigger coverage while reducing the test generation time. 
Conceptually similar to the \textit{`N-detection test'} in stuck-at automatic test pattern generation (ATPG), MERO~\cite{chakraborty2009mero}
leverages random test vectors and mutates them until each rare node in the circuit is individually triggered at least $N$ times. 
Such a simple detection heuristic results in an unsatisfying trigger coverage, particularly Trojans that are hard-to-activate. 
To overcome the limitation of MERO,~\cite{saha2015improved} proposes to use genetic algorithms (GA) and Boolean Satisfiability (SAT) to produce test inputs that excite regular rare nodes and internal \textit{hard-to-trigger} nodes, respectively.
As the end result,~\cite{saha2015improved} achieves a higher trigger coverage compared to MERO, while it is {inefficient} due to the long test generation time. 
TRIAGE~\cite{nourian2018hardware} further improves GA-based test generation by devising a more appropriate `fitness' function that incorporates the controllability and observability factors of rare nodes.
{However, the GA nature of TRIAGE limits its efficiency for test input space exploration and the resulting test set might be unnecessarily large. }
{TGRL~\cite{pan2021automated} suggests to train a machine learning model for test patterns generation that combines rare signal stimulation as well as controllability/observability analysis. Although TGRL claims to explore reinforcement learning, its test pattern generation pipeline (Alg.3 in~\cite{pan2021automated}) does not involve sequential decision making in standard RL techniques. Instead, TGRL learns a ML model via stochastic gradient descent for TPG.}
\vspace{-0.6em}
\subsection{Reinforcement Learning} \label{sec:rl_basics}
\begin{figure}[ht!]
    \centering
    \includegraphics[width=0.45\textwidth]{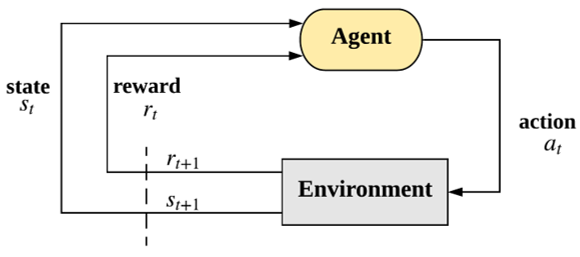} 
    \vspace{-1.5em}
    \caption{Illustration of the agent-environment interaction in reinforcement learning.}
    \label{fig:RL_demo}
    \vspace{-1em}
\end{figure}

\vspace{-0.4em}
Reinforcement learning~\cite{kaelbling1996reinforcement,wiering2012reinforcement,sutton2018reinforcement} is a machine learning technique that is capable of solving complex problems in various domains. 
RL works \textit{sequentially} in an environment by taking an action, evaluating its reward, and adjusting the following actions accordingly. 
In particular, an RL paradigm involves an \textit{agent} that observes the environment and takes \textit{actions} to maximize the \textit{reward} determined by the problem of concern~\cite{sutton2018reinforcement,mnih2013playing}.
Figure~\ref{fig:RL_demo} shows the interaction between the agent and the environment in the RL paradigm.

\vspace{-1.5em}

We introduce the key concepts in an RL system below:

\vspace{0.2em}
{\tikz\draw[black,fill=black] (-0.5em,-0.5em) rectangle (-0.2em,-0.2em);} \textbf{Action Space.} 
The action space is a set of possible moves that the agent can take to change to a new state.
For example, in a video game, an action can be running left/right, or jumping high/low. 

\vspace{0.2em}
{\tikz\draw[black,fill=black] (-0.5em,-0.5em) rectangle (-0.2em,-0.2em);} \textbf{Environment.}
The environment takes the agent’s current state and action as input, and returns the reward and the next state as the output. 
Depending on the problem domain, the environment might be a set of physical laws or chemical reaction rules that processes the actions and establish the corresponding outcomes.

\vspace{0.2em}
{\tikz\draw[black,fill=black] (-0.5em,-0.5em) rectangle (-0.2em,-0.2em);} \textbf{State.} 
A state is a concrete and instantaneous situation in which the agent finds itself.
This can be an instant configuration, a particular place and a moment that puts the agent in connection with other influential objects in the environment, such as opponents or awards.
It is noteworthy that a state needs to contain all information to ensure the system satisfies the \textit{Markov property}~\cite{pardo2018time}. 

\vspace{0.2em}
{\tikz\draw[black,fill=black] (-0.5em,-0.5em) rectangle (-0.2em,-0.2em);} \textbf{Observations.} 
The agent can obtain observations (emission of states) from the environment.  
In particular, the observation is a (stochastic) function of the state. 

\vspace{0.2em}
{\tikz\draw[black,fill=black] (-0.5em,-0.5em) rectangle (-0.2em,-0.2em);} \textbf{Reward.} 
The reward is a numerical value that evaluates the fitness (success or failure) of an agent’s actions in a given state. 
From a given state, an agent takes actions in the environment and acquires the new state as well as the reward from the environment.
A \textit{cumulative reward} is defined as the summation of discounted rewards: $G(t)=\sum_{k=0}^{n} \gamma^{k} R(t+k+1)$. 
The discount factor $\gamma$ ($0 \leq \gamma \leq 1$) tunes the importance of future rewards for the current state.  
The key idea of RL is to find a series of actions that maximize the expected cumulative reward. 

\vspace{0.2em}
{\tikz\draw[black,fill=black] (-0.5em,-0.5em) rectangle (-0.2em,-0.2em);}  
{\textbf{Policy.} The policy of a RL algorithm is typically defined within the context of Markov decision process~\cite{sutton2018reinforcement}. Given the state information, policy is the suggested action that the agent shall take in order to obtain a high reward.}

Our objective is to develop an adaptive test pattern generation framework for \textit{logic testing} with high Trojan coverage and small test set size. {Therefore, \sys{} belongs to the test-time detection category introduced in Section~\ref{sec:HT_detect}.
We choose RL over other machine learning techniques (e.g. neural networks) since the reward-oriented and progressive nature of RL makes it appealing for our goal.}
Furthermore, to reduce the complexity of RL, \sys{} integrates adaptive sampling to prioritize test patterns that provide more useful information for HT detection. 
\vspace{-0.5em}
\section{\sys{} Overview}  \label{sec:overview}
\vspace{-0.6em}
In this section, we first discuss the limitations of prior works on Hardware Trojan detection and our motivation (Section~\ref{sec:motiv}), then introduce our assumptions and threat model for \sys{} framework (Section~\ref{sec:threat}). 
We demonstrate the overall workflow of \sys{} test pattern generation technique in Section~\ref{sec:global}. 
\sys{} is a hardware-friendly framework and we present our architecture design in Section~\ref{sec:hardware}.

\vspace{-0.8em}
\subsection{Motivation and Challenges}   \label{sec:motiv}
\vspace{-0.5em}
Prior works have advanced logic testing-based Trojan detection using various techniques~\cite{chakraborty2009mero, saha2015improved, nourian2018hardware}.  
We discuss the limitations of these 
detection schemes below.

\vspace{0.1em}
\noindent \textbf{MERO.} Inspired by the traditional `N-detect' test used in stuck-at ATPG, MERO~\cite{chakraborty2009mero} generates random test vectors to activate each rare node (identified as nodes with transition probability smaller than the threshold $\theta$ ) to the corresponding rare value at least $N$ times. 
MERO has three main disadvantages: 
(i) Triggering all rare nodes for $N$ times might be very time-consuming or even impractical; 
(ii) It yields low trigger coverage for hard-to-trigger Trojans;
(iii) It only explores a small number of test vectors in the entire possible space {due to its bit mutation and test vector selection policy.}

\vspace{0.1em}
\noindent \textbf{ATPG based on GA+SAT.} The paper~\cite{saha2015improved} combines genetic algorithms and SAT in test pattern generation for HT detection. While it improves the trigger coverage compared to MERO,~\cite{saha2015improved} has two constraints: slow test set generation and large memory footprint.

\vspace{0.1em}
\noindent \textbf{TRIAGE.} The paper~\cite{nourian2018hardware} proposes TRIAGE that integrates the benefits of MERO and~\cite{saha2015improved}. 
TRIAGE leverages the SCOAP testability parameters and advises the fitness function of GA for HT detection. 
{However, the evolutionary nature of GA determines that TRIAGE might be `trapped' in the vicinity of a local optimum, thus exploring only a small portion of the full test input space. }


We present \sys{} as a holistic solution to address the limitations of the previous works. 
To this end, we identify three main challenges of developing an efficient and effective logic testing-based HT detection technique as follows:

\vspace{0.1em}
\noindent \textbf{(C1) High trigger coverage.} The test vector set shall yield a high trigger coverage rate to ensure that the probability of activating the stealthy Trojan is large. This property is critical for the \textit{effectiveness} criterion of HT detection.

\vspace{0.1em}
\noindent \textbf{(C2) Efficient test generation.} The runtime overhead of test pattern generation shall be reasonable while attaining a high trigger coverage. 
For hardware-assisted security, this implies that a test set with smaller size is preferred. 
This requirement assures the efficiency and practicality of the HT detection method, particularly on large circuits.  

\vspace{0.1em}
\noindent \textbf{(C3) Scalable to large benchmarks.} 
The runtime consumed by the test pattern generation technique shall not scale exponentially with the size of the examined circuit. 

\sys{} tackles the above challenges $(C1) \sim (C3)$ using an \textit{adaptive, RL-based} input space exploration approach. 
Furthermore, we provide architecture design for \sys{}-based TPG in Section~\ref{sec:hardware} to enable hardware-assisted security.
{We empirically corroborate the superior performance of \sys{} compared to the above counterparts in Section~\ref{sec:eval}.}

\vspace{-0.5em}
\subsection{Threat Model} \label{sec:threat}
\vspace{-0.5em}
As shown in Figure~\ref{fig:HT_demo}, HTs consist of two parts: trigger and payload. 
Figure~\ref{fig:HT_demo} shows an example of HT design.
\sys{} is applicable to both combinational and sequential circuits. 
{One can unroll sequential circuits into combinational ones and apply \sys{} for test pattern generation.}
Without the loss of generality, we assume that the adversary uses a logic-AND gate as the Trojan trigger that takes a subset of rare nodes as its inputs. 
An XOR gate is used to flip the value of the payload node when the trigger is activated (i.e., each of the trigger nodes has a logical value `1'). 

We make the following assumptions about \sys{} framework:

\vspace{0.2em}
\noindent \textbf{(i) The defender knows the netlist of the circuit under test.} 
We assume the party that executes logic testing has the netlist description of the circuit to be examined. 
This netlist can be obtained by performing de-packaging, de-layering, and imaging~\cite{torrance2009state,meade2016netlist,li2012reverse,fyrbiak2017hardware} on the physical circuit. 
{While hardware obfuscation techniques such as camouflaging~\cite{li2017provably,yasin2016camoperturb,shakya2019covert,shamsi2019impossibility} and logic encryption~\cite{yasin2015improving,yasin2017evolution,xie2018anti,tan2020benchmarking} could make the trigger design of the Trojan harder to identify, we consider the scenario where the circuit under test is not encrypted in our threat model since this setting is also used in previous Trojan detection papers~\cite{chakraborty2009mero,pan2021automated,shakya2017benchmarking,yang2020survey}. }

\vspace{0.2em}
\noindent \textbf{(ii) The defender can observe the `indication signal' when the Trojan is activated.} 
We assume the defender can observe certain \textit{manifestations} of the hidden Trojan when it is activated. 
In particular, we assume the defender knows the correct response of the CUT to a given test input and observes the primary outputs of the CUT for comparison. 
Note that \sys{} is compatible with techniques that increase manifestation signals (e.g., test point insertion).

\vspace{-0.3em}
\subsection{Global Flow}  \label{sec:global}
\begin{figure*}[ht!]
\centering
 \includegraphics[width=0.99\textwidth]{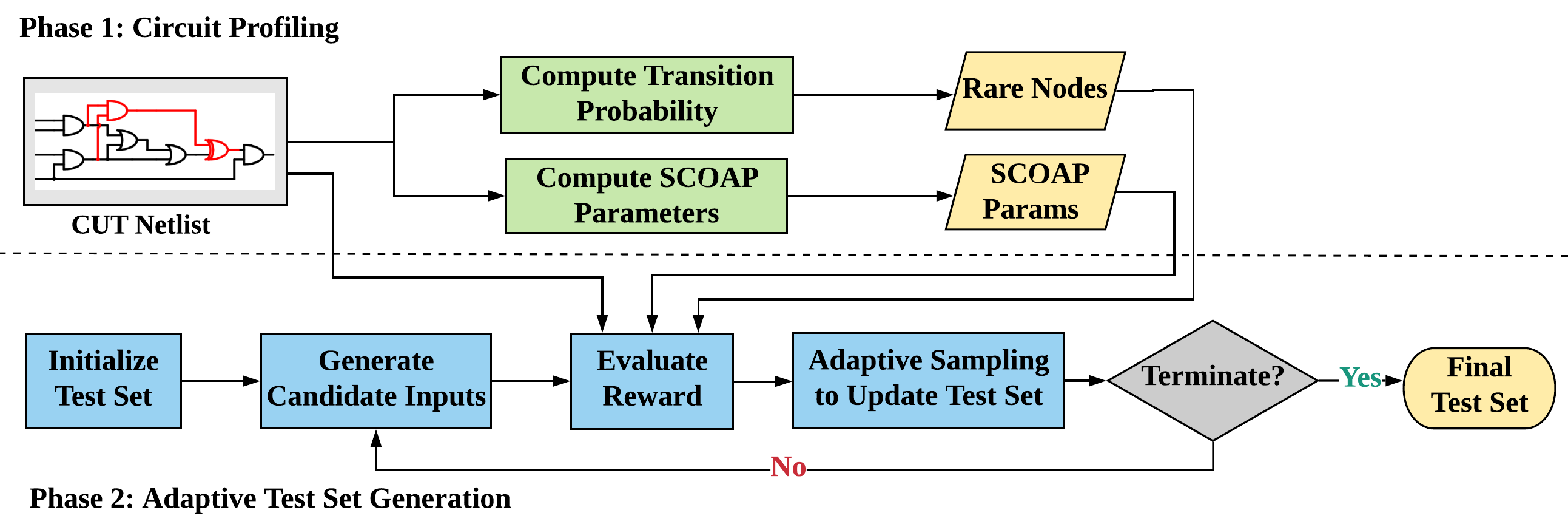} 
 \vspace{-0.3em}
\caption{\label{fig:global} Global flow of \sys{} framework for Hardware Trojan detection. }
\end{figure*}

\vspace{-0.3em}
Figure~\ref{fig:global} illustrates the global flow of \sys{}. 
We discuss the threat model in Section~\ref{sec:threat}.
\sys{} framework consists of two stages: (i) Circuit profiling phase (offline) that computes the transition probabilities and SCOAP testability parameters of the netlist;
(ii) Adaptive RL-based test set generation phase (online) that progressively identifies test vectors with high reward values. 

\setlength{\belowcaptionskip}{-6pt} 
\vspace{-0.6em}

\vspace{0.4em}
\noindent \textbf{Phase I: Circuit Profiling.} This stage includes the following:

\textbf{(1) Compute Transition Probabilities.}
Given the netlist of the circuit under test, \sys{} first computes the \textit{transition probability} of each internal node in the netlist. 
In particular, we use the method in~\cite{salmani2011novel} and assume that each primary input has an equal probability of taking a logical value of 0 and 1. {We make this assumption about the primary input values since previous Trojan detection papers~\cite{salmani2011novel, xiao2016hardware, bhunia2014hardware,li2016survey} use the same assumption when computing the transition probability.}
Mathematically, the transition probability of a node is computed as $P_{trans}=p(1-p)$ where $p=Prob(node=1)$.
$P_{trans}$ of each node is then compared with a pre-defined threshold $\theta$ to identify the \textit{rare nodes}. 
{Identifying rare nodes is important for HT detection since the defender does not know the exact set of trigger nodes used by the attacker. As such, the activation status of rare nodes provides guidance to generate test inputs that are likely to trigger the stealthy Trojan.}

\vspace{0.2em}
\textbf{(2) Compute SCOAP Testability Parameters.}
Controllability and observability are important testability characteristics of a digital circuit.
More specifically, \textit{`controllability'} describes the ability to establish a specific node to 0 or 1 by setting the primary inputs.  
\textit{`Observability'} defines the capability of determining the value of a node by controlling the circuit's inputs and observing the outputs. 
The \textit{testability} parameters are useful for Trojan detection since they allow \sys{} to distinguish the quality of different rare nodes.

\vspace{0.3em}
\noindent \textbf{Phase II: Adaptive RL-based test pattern generation.} 
After the CUT is profiled offline in Phase 1, \sys{} performs adaptive test input generation as shown in the bottom of Figure~\ref{fig:global}. 
We outline each step as follows:  

\vspace{0.2em}
\textbf{(1) Initialize Test Set.} \sys{} first generates an initial test vector set that is used in the later steps.
A naive way to do so is random initialization, which may not be optimal for HT detection.
To improve the trigger coverage in the later runs, \sys{} employs SAT to find a number of test inputs that activate a subset of rare nodes. We call this method \textit{`smart initialization'} and empirically corroborate its effectiveness in Section~\ref{sec:effect}. 

\vspace{0.2em}
\textbf{(2) Generate Candidate Test Inputs.} In each iteration of \sys{}'s adaptive test vector generation, we first produce a sufficient number of candidate test input patterns that might improve the detection performance when added to the current test set. \sys{} deploys random test generation for this purpose.

\vspace{0.2em}
\textbf{(3) Evaluate Reward Function.} \sys{} applies the candidate test inputs on the examined circuit and collects the observations, i.e., the netlist status represented as a directed acyclic graph (DAG). We incorporate the transition probabilities and the SCOAP testability parameters from Phase 1 as well as a novel DAG-level diversity measure to define our reward function. 

\textbf{(4) Adaptive Sampling to Update Test Set.} Inspired by the selection step in genetic algorithms, we design an adaptive sampling module that picks `high-quality' test patterns for fast and efficient input space exploration.  
In particular, after computing the reward value of each test input in the candidate test vectors, \sys{} selects the ones with the highest scores and append them to the current test set. 

At the end of each iteration, \sys{} checks the termination condition and decides whether or not the progressive test generation process shall continue.

\vspace{0.2em}
\noindent \textbf{Performance Metrics.} 
We use \textit{effectiveness} and \textit{efficiency} as two main metrics to assess the performance of a Trojan detection scheme.
In particular, we measure the effectiveness from two aspects: trigger coverage and Trojan coverage (i.e. detection rate). 
The efficiency property is measured by the test set generation time and test set size.  
\sys{}, for the first time, provides the trade-off between effectiveness and efficiency by adaptively generating a set of test patterns with evolving quality over time.
The quantitative analysis of the above metrics is demonstrated in Section~\ref{sec:eval}.

\vspace{-0.3em}
\section{\sys{} Algorithm Design}  \label{sec:alg}
\vspace{-0.5em}
The key to ensuring a high probability of Trojan detection using logic testing is to generate a test set that can trigger the circuit to diverse states, in particular, the rare nodes in the circuit. 
To this end, \sys{} leverages three important characteristics of the circuit: the transition probabilities, the SCOAP testability measures, and the DAG-level diversity. 
In particular, \sys{} employs an \textit{RL-driven} test pattern generation approach that uses the above three properties to progressively generate test inputs.  
{Inspired by the selection stage in genetic algorithms, we integrate an adaptive sampling module that progressively expands the current test set (used as historical information) with high-quality test patterns. This \textbf{response-adaptive} design is beneficial for statistical search of the HT trigger in the circuit input space, thus improves the efficiency of \sys{}'s RL-based pipeline.}
We detail the two main phases of \sys{} shown in Figure~\ref{fig:global} in the following of this section.

\vspace{-0.6em}
\subsection{Circuit Profiling} \label{sec:circuit_profile}
\vspace{-0.3em}
Alg.~\ref{alg:circuit_profile} outlines the steps of the circuit profiling phase in \sys{}.
This stage obtains two informative properties of the circuit: the transition probabilities and testability measures. 
In particular, we use \textit{random testing} and \textit{logic simulation} to estimate the transition probability $P_{trans}$ of each node in the netlist $C_n$. 
To further investigate the rewards of different rare nodes, \sys{} also computes the SCOAP parameters of the nodes using the technique in~\cite{goldstein1980scoap}. 

\sys{}'s circuit profiling stage characterizes the \textit{static reward} properties of the circuit in terms of the transition probabilities of rare nodes and testability measures. 
We call these two properties \textit{`static'} since they are \textit{independent} of the circuit input for a given circuit netlist. 
As such, our profiling phase can be performed \textit{offline}. 
The above two properties are indispensable for the reward computation step in Phase 2 of \sys{} since:
(i) Transition probabilities and rare nodes shed light on the potential trigger nodes exploited by the malicious adversary. The defender knows that a subset of rare nodes are used to design the stealthy Trojan while he has no knowledge about the exact trigger set. As such, rewarding the activation of rare nodes encourages the test vectors to stimulate the possible HT. {(Note that the Trojan activation condition is equivalent to knowledge of the exact trigger set and both are assumed to be unknown to the defender.)}  
(ii) Testability parameters provide more fine-grained information about the quality of individual rare nodes in the context of HT detection.  
One can compare the fitness of two test inputs by counting and comparing the number of activated rare nodes correspond to each test vector. 
However, such a naive counting mechanism neglects the intrinsic difference between the quality of individual rare nodes. In principle, a rare node with higher controllability and observability shall be assigned with higher reward values. 
As such, \sys{} integrates the SCOAP testability measures to quantify the reward of each activated rare node. 

\algnewcommand\algorithmicinput{\textbf{INPUT:}}
\algnewcommand\INPUT{\item[\algorithmicinput]}
\algnewcommand\algorithmicoutput{\textbf{OUTPUT:}}
\algnewcommand\OUTPUT{\item[\algorithmicoutput]}

\vspace{-0.3em}
\begin{algorithm}[ht!]
\caption{Circuit Profiling. }
\label{alg:circuit_profile}
\begin{algorithmic}[1]
\INPUT \textbf{Netlist of the circuit under test ($C_n$); Number of random tests ($H$); Threshold on transition probability ($\theta$) for rare nodes.}

\vspace{0.5em}
\OUTPUT \textbf{The set of rare nodes ($R$); Computed testability parameters $TP$ = ($CC0, CC1, CO$).}

\vspace{0.2em}
\State Initialize rare node set: $R \gets \emptyset$
\vspace{0.2em}
\State Generate random inputs: $I \gets RandGen(C_n,~H)$.
\vspace{0.2em}
\State Perform logic simulation: $O \gets LogicSim(C_n,~I)$.
\vspace{0.2em}

\For{node in $C_n$}
    \State Compute frequency: $p= CountOnes(O, node)/H$
    \vspace{0.2em}
    \State Estimate transition probability: $P_{trans} = p(1-p)$
    \vspace{0.2em}
    \If{$P_{trans} < \theta$}
        \State $R \gets R \cup node$
    \EndIf
\EndFor

\State Obtain SCOAP parameters: 
    \vspace{0.2em}
    \Statex \hspace{\algorithmicindent} $~(CC0,~CC1,~CO) \gets ComputeSCOAP(C_n)$
\\
\vspace{0.2em}
\noindent \textbf{Return:} Obtained rare node set $R$, SCOAP testability parameters $TP=(CC0, CC1, CO)$.
\end{algorithmic}
\end{algorithm}

\subsection{Adaptive RL-based Test Pattern Generation} \label{sec:adaptive_rl}
\vspace{-0.5em}
\sys{} deploys a \textit{progressive, reinforcement learning-driven} algorithm for efficient and effective test input space exploration with the goal of HT detection.
Section~\ref{sec:rl_basics} introduces the basic concepts of RL. 
We discuss how we map the Trojan detection problem to the RL paradigm as follows. 

\vspace{0.1em}
\noindent \textbf{\sys{}'s RL Formulation of Trojan Detection:}

\vspace{0.1em}
{\tikz\draw[black,fill=black] (-0.5em,-0.5em) rectangle (-0.2em,-0.2em);} \textbf{State.}
The objective of \sys{} is to adaptively generate test patterns with high effectiveness for Trojan detection in an iterative manner. 
As such, \sys{} defines a \textit{state} as the \textit{current test set} in the present iteration. 

\vspace{0.1em}
{\tikz\draw[black,fill=black] (-0.5em,-0.5em) rectangle (-0.2em,-0.2em);} \textbf{Action Space.} 
Recall that an action transforms the agent into a new state, which is the new test set according to our definition of the state above. 
Therefore, a feasible \textit{action} for \sys{} is to \textit{identify a set of new test input vectors} in each iteration that improves the quality of HT detection when added to the current test set. 

\vspace{0.1em}
{\tikz\draw[black,fill=black] (-0.5em,-0.5em) rectangle (-0.2em,-0.2em);} \textbf{Environment.} For HT detection, the \textit{netlist of the circuit} ($C_n$) can be considered as the \textit{environment} that converts the current state and the action, and returns the reward value. 

\vspace{0.1em}
{\tikz\draw[black,fill=black] (-0.5em,-0.5em) rectangle (-0.2em,-0.2em);} \textbf{Observations.} 
The agent makes the observation of the environment before reward computation. 
For Trojan detection problems, we model the \textit{DAG formed by the values of all nodes} in the netlist given a specific input vector as an \textit{observation} of the circuit state. 
  
\vspace{0.1em}
{\tikz\draw[black,fill=black] (-0.5em,-0.5em) rectangle (-0.2em,-0.2em);} \textbf{Reward.}
The definition of the reward function directly reflects the objective of the problem that one aims to solve. 
As such, for the task of logic testing-based HT detection, \sys{} designs a \textit{composite reward} function to encourage the generation/exploration of test inputs that facilitate the excitation of the potential HT.

The mathematical definition of \sys{}'s \textit{dynamic} reward function is given in the equation below:
\vspace{-0.2em}
\begin{align}
    \label{eq:reward}
    Reward(T_i|~S_i) &= \lambda_1 \cdot V_{rare}(T_i, ~R) + \lambda_2 \cdot V_{scoap}(T_i, ~R, ~TP) \nonumber \\
    &+ \lambda_3 \cdot V_{DAG}(T_i| ~S_i).    
\end{align}
Here, $S_i$ and $T_i$ are the current test set (i.e., the state) and the newly generated test inputs in $i^{th}$ iteration, respectively.  
$R$ and $TP$ are the set of rare nodes and the SCOAP testability parameters identified in Phase 1 (\textit{static attributes}). 
The hyper-parameters $\lambda_1$, $\lambda_2$, $\lambda_3$ determine the relative weighting of the three reward terms.   
The reward function $Reward(T_i|~S_i)$ characterizes the fitness of the specific test inputs $T_i$ while considering the current test set $S_i$. 
Evaluating the reward value of $T_i$ \textit{in the context of the historical test patterns} ($S_i$) makes \sys{}'s RL framework \textit{adaptive} and intelligent. 


We detail how each term in \sys{}'s reward function is designed below.
Inspired by the `N-detect' test, the first reward term in Equation~(\ref{eq:reward}) aims to activate each rare node in the circuit for at least $N$ times. 
To this end, we define the \textbf{rare node reward} $R_{rare}$ as follows:
\begin{equation}
    \label{eq:reward_rare}
    \vspace{-0.4em}
    V_{rare}(T_i, ~R) = -\sum_{r \in R} ~abs(N-Ctr_i(r)), 
    \vspace{-0.4em}
\end{equation}
where $Ctr_i(r)$ is the number of times that the rare node $r$ is activated to its rare value up to the $i^{th}$ iteration.

The second reward term in Equation~(\ref{eq:reward}) leverages the SCOAP parameter $TP=(CC0, CC1, CO)$ computed in Phase 1 to encourage the stimulation of rare nodes with high controllability and observability. 
Given the current test set $S_i$, we can obtain the set of activated rare nodes $Rtr_i$ (which is a subset of $R$). 
The \textbf{SCOAP testability reward} $V_{scoap}$ is then computed as follows: 
\begin{equation}
    \label{eq:reward_scoap}
    V_{scoap}(T_i,~R,~TP) = \sum_{r \in Rtr_i} CC(r) + CO(r).
\end{equation}
Here, $CC(r)$ denotes the controllability of setting the rare node $r$ to its corresponding rare value. 
More specifically, $CC(r)$ shall be converted to $CC0(r)$ or $CC1(r)$ depending on the rare value of the node $r$. 
$CO(r)$ denotes the observability of the rare node $r$.

Besides leveraging the static attributes identified in Phase 1 to define the rare node reward $R_{rare}$ and the SCOAP testability reward $R_{scoap}$, \sys{} further explores the \textbf{graph-level diversity} extracted from the circuit netlist. 
In particular, \sys{} identifies the {dynamic fitness property}, i.e., the {DAG-level diversity} that is jointly determined by the circuit netlist and the test vector set. 
Such a DAG-level distance serves as a \textit{dynamic} fitness measure since it is \textit{input-aware}. 
Recall that \sys{} leverages an RL paradigm and considers the value assignments of all nodes when given the netlist $C_n$ and a specific test input as the observation. 
We use the \textit{graph representation} of the circuit to abstract the observed netlist status. To facilitate the computation, \sys{} flattens the DAG to an \textit{ordered sequence} based on the circuit level information. 
The distance between the two transformed DAG sequences is used as the DAG-level diversity measure. 
To summarize, we define the \textbf{DAG diversity reward} as follows: 
\vspace{-0.3em}
\begin{equation}
    \label{eq:reward_dag}
    \scalebox{0.96}{
    $V_{DAG}(Ti|~S_i; C_n) = HammDist(DAG(T_i;~C_n),~DAG(S_i;~C_n)).$
    }
    \vspace{-0.3em}
\end{equation}
Here, $DAG(T_i; C_n)$ denotes the flattened ordered sequence of the DAG obtained when applying the test inputs $T_i$ to the circuit $C_n$. 
{The diversity measurement function $HammDist$ computes the normalized pairwise distance of the flattened DAGs using the Hamming distance metric.} 
Since the DAG sequence of the circuit is binary-valued (0 or 1), \sys{} employs $XOR$ function as an efficient implementation of the $HammDist$ function.    
It's worth noting that this graph reward $V_{DAG}$ is aware of historical test inputs ($S_i$), thus provides guidance to select new inputs that stimulate different internal nodes structure in the context of current test inputs $S_i$.

\vspace{0.1em}
{\tikz\draw[black,fill=black] (-0.5em,-0.5em) rectangle (-0.2em,-0.2em);} {\textbf{Policy.} The policy component of a RL algorithm suggests actions to achieve a high reward given the current state. Recall that \sys{} defines the state and the action space as the current set of test vectors and the expansion with the new test patterns, respectively. Therefore, the policy module of \sys{} selects the most suitable test pattern candidates and add them to the result test set (line 5\&6 in Alg.~\ref{alg:adaptive_rl}).}

Algorithm~\ref{alg:adaptive_rl} outlines the procedure of our adaptive test set generation framework. 
{We emphasize that \textbf{\sys{} does not require explicit training} on the training set, which is typically required by machine learning model (e.g., gradient descent-based training). 
The RL nature enables \sys{} to search for distinguishing test inputs with the guidance of the composite reward.} {This makes our detection method fundamentally different from TGRL~\cite{pan2021automated} that still trains a ML model for test pattern generation.}
We discuss how \sys{} leverages the RL paradigm formulated above to achieve logic testing-based HT detection in the following of this section.

\setlength{\textfloatsep}{5pt}
\begin{algorithm}[ht!]
\caption{Adaptive Reinforcement Learning based Test Input Pattern Generation.}
\label{alg:adaptive_rl}
\begin{algorithmic}[1]
\INPUT \textbf{Netlist of circuit under test ($C_n$); Rare node set $R$; SCOAP testability parameters $TP=(CC0, CC1, CO)$; Size of candidate test inputs per iteration ($M$); Size of selected test inputs per iteration ($L$); Maximal number of iterations ($I_{max}$); 
Percentage threshold of rare nodes ($p$); Target activation times ($N$). } 

\vspace{0.5em}
\OUTPUT \textbf{A set of test patterns $S$ for Trojan detection of the target circuit $C_n$. }

\vspace{0.2em}
\State Initialization: 
    \vspace{0.2em}
    \Statex \hspace{\algorithmicindent} $ S_0 = \left\{\vec{S}_0^1, ..., \vec{S}_0^L \right\} \gets SmartInitialize(L)$. 

    \Statex \hspace{\algorithmicindent} Iteration counter: $i \gets 0$  
  
\While{$ i < I_{max}$ and HT is not activated}
    \vspace{0.2em}
    \State $ T_i \gets GenerateTestCandidates(M;~C_n)$ 
    \vspace{0.2em}
    \State $Reward(T_i|~S_i) \gets EvaluateReward(T_i, ~S_i; ~C_n)$
    \vspace{0.2em}
    \State $T^{top}_i \gets SelectTopCandidates(T_i,~Reward,~L)$

    \vspace{0.2em}
    \State Update test set: $S_{i+1} \gets S_i \cup T^{top}_i $   \Comment{Adaptive sampling to expand test set}

    \vspace{0.2em}
    \State $A_i \gets CountRareNodeActivation(S_i; ~C_n)$
    
    \vspace{0.2em}
    \If{$p\%$ elements in $A_i$ $\geq N$ \& $A_i.min() \geq 1$}  \Comment{Check termination condition} 
        \State break   
    \EndIf
    \vspace{0.2em}
    \State $i \gets i+1$
\EndWhile
\\
\noindent \textbf{Return:} Obtained a test set ($S_i$) for logic testing-based HT detection of the circuit $C_n$.
\end{algorithmic}
\end{algorithm}

\vspace{-0.2em}
\noindent \textbf{\tikz\draw[black, fill=black] (0,0) circle (0.93ex) node[white] {1};} \textbf{Smart Initialization.}
Recall that the intuition of logic testing-based Trojan detection is to encourage the generation of test inputs that activate diverse combinations of rare nodes to their corresponding rare values. 
Random test vectors might be unlikely to yield a high trigger coverage, especially on large circuits. 
To explore the above intuition, \sys{} leverages SAT to generate the initial test set (line 1 in Algorithm~\ref{alg:adaptive_rl}) such that it is able to activate diverse rare nodes specified by the defender. 
We empirically validate the advantage of our smart initialization as opposed to the random variant in Section~\ref{sec:effect}. 
{It is worth noticing that while the defender can identify rare nodes in the circuit by thresholding the transition probabilities, it might be unfeasible to find an input that stimulates all rare nodes to their rare values. Therefore, \sys{} tries to generate test patterns that stimulate different combinations of rare nodes for Trojan detection.}

\noindent \textbf{\tikz\draw[black, fill=black] (0,0) circle (0.93ex) node[white] {2};} \textbf{Generate Candidate Test Patterns.}
\sys{} progressively identifies test inputs that are suitable for HT detection using an iterative approach. 
To this end, \sys{} first generates a sufficient number of candidate test vectors at the beginning of each iteration (line 3 in Alg.~\ref{alg:adaptive_rl}). 
These candidates are responsible for exploring the test input space and aim to find solutions with high rewards.  
In our experiments, we adopt an adaptive sampling method to generate candidate test patterns at each iteration. In particular, the sampling weights for the test vectors in the initial set $S_0$ are uniformly assigned at iteration 0. In other words, at iteration 0, we perform a uniform sampling to generate candidate test patterns. Then the sampling weights of test vectors at iteration $i+1$ will be updated based on the normalized reward values evaluated at iteration $i$. Test vectors with higher reward values will result in higher sampling weights, which in turn increases the probability of the test vectors to be included in the generated set $S$. The adaptive sampling method allows us to optimize test pattern generation by favoring test patterns with higher reward values thus enhance convergence in our test pattern generation.    


\vspace{-0.0em}
\noindent \textbf{\tikz\draw[black, fill=black] (0,0) circle (0.93ex) node[white] {3};} \textbf{Evaluate Reward Function.} 
The definition of reward is task-specific. 
Since our objective is to generate test patterns that stimulate the circuit (particularly the rare nodes) to different states for Trojan detection, \sys{} designs an innovative composite reward function as shown in Equation~(\ref{eq:reward}). 
In each iteration, the reward values of the candidate test inputs are evaluated (line 4 of Alg.~\ref{alg:adaptive_rl}). 
Our compound reward function captures informative features that are beneficial for HT detection from three aspects: the number of times that each rare node is activated ($V_{rare}$), the SCOAP testability measures that quantify the fitness of different rare nodes ($V_{scoap}$), and the graph-level diversity between the current test inputs and historical ones ($V_{DAG}$). 

\vspace{-0.2em}
\noindent \textbf{\tikz\draw[black, fill=black] (0,0) circle (0.93ex) node[white] {4};} \textbf{Adaptive Sampling to Update Test Set.} Recall that in \sys{}'s RL paradigm, the current test set $S_i$ represents the `state' variable. 
After obtaining the reward values of individual candidate test input in $T_i$ from Step 3, \sys{} updates the state by selecting a subset of $T_i$ that has the highest reward values and adding them to the current test set $S_i$. 
This step is conceptually similar to the selection stage in genetic algorithms. With the domain-specific definition of reward, \sys{} adaptively samples high-quality test patterns from the randomly generated candidate test inputs, therefore facilitates fast exploration of the circuit input space for HT detection.

\vspace{-0.2em}
\noindent \textbf{\tikz\draw[black, fill=black] (0,0) circle (0.93ex) node[white] {5};} \textbf{Check Termination Condition.} 
\sys{}'s adaptive test set generation terminates if any of the following three conditions is satisfied: (i) $p\%$ of all rare nodes are activated for at least $N$ times and all rare nodes are activated at lease once (line 8 in Alg.~\ref{alg:adaptive_rl});
(ii) The maximal number of iteration $I_{max}$ is reached (line 2 in Alg.~\ref{alg:adaptive_rl}); 
(iii) The current test set $S_i$ activates the hidden Trojan, i.e., all involved trigger nodes are activated to their corresponding rare values by $S_i$ (line 2 in Alg.~\ref{alg:adaptive_rl}). 
Note that we include termination condition (iii) since our threat model assumes that the defender can observe the manifestation of an activated Trojan. 

\noindent \textbf{Discussion.} {As summarized in Alg.~\ref{alg:adaptive_rl}, our reinforcement learning approach does not require model training. Instead, we progressively generate the set of test vectors using adaptive sampling given the particular circuit with the goal of maximizing the RL rewards for Trojan detection. From this perspective, our RL-based detection tool generates a specific test set for the circuit under test. 
However, \sys{} is generic in the sense that it is agnostic to the circuit structure and can be applied to other different circuits (i.e., re-applying \sys{} to other circuits does not require any model training since we do not incorporate neural networks in our RL detection pipeline shown in Alg.~\ref{alg:adaptive_rl}). }
\vspace{-0.5em}
\section{\sys{} Architecture Design}  \label{sec:hardware}
\vspace{-0.8em}
Beyond the novel test generation algorithm discussed in Section~\ref{sec:alg}, we design a Domain-specific systems-on-chip (DSSoC) architecture of \sys{} for its practical deployment. The bottleneck of \sys{} implementation is the computation of the test input's reward $Reward(T_i | S_i)$ according to Equation~(\ref{eq:reward}). 
Given the rare node set $R$ and SCOAP testability measures of the circuit $TP$ from offline circuit profiling (Algorithm~\ref{alg:circuit_profile}), the online reward evaluation of a new test input $T_i$ involves three terms as shown in Equation~(\ref{eq:reward}): identifying the rare nodes stimulated by $T_i$ (for $V_{rare}$), obtaining the SCOAP values corresponding to each active rare node (for $V_{scoap}$), and computing the DAG-level graph distance (for $V_{DAG}$). 
{Note that the third component require us to obtain the DAG with nodes value assignment when applying the test input on the circuit $DAG(T_i;~C_n)$. This information is also sufficient to compute the first two reward terms.} Therefore, the main task for \sys{}'s on-chip implementation is to obtain the value-assigned DAG for a new test input on the circuit ($DAG(T_i;~C_n)$).

To accelerate circuit evaluation, \sys{} deploys \textit{circuit emulation} on the programmable hardware to obtain the response $DAG(T_i;~C_n)$. Furthermore, \sys{} constructs the customized auxiliary circuitry automatically to pipeline each computation stage and reduce the runtime overhead. 
We design an optimized DSSoC architecture of \sys{} for efficient implementation of our adaptive TPG method outlined in Algorithm~\ref{alg:adaptive_rl}.

\vspace{-0.6em}
\subsection{Architecture Overview} \label{sec:arch_overview}
\begin{figure*}[ht!]
\centering
 \includegraphics[width=0.98\textwidth]{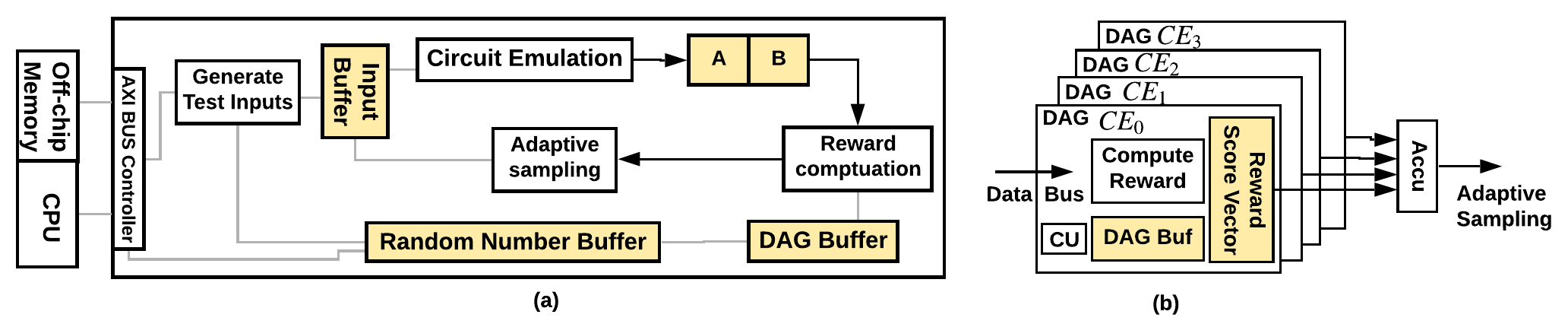} 
 \vspace{-0.4em}
\caption{\label{fig:arch} Overview of \sys{} architecture design. The overall layout of the hardware system (a) and the implementation of Reward Computation Engines (b) are shown. }
\end{figure*}

\vspace{-0.6em}
The overall hardware architecture of \sys{}'s online test patterns generation is shown in Figure~\ref{fig:arch} (a). 
\sys{} leverages Algorithm/Software/Hardware co-design approach to accelerate the test inputs searching process shown in Figure~\ref{fig:global} (phase2). More specifically, \sys{} maps the netlist of the circuit under test ($C_n$) with the auxiliary part to the FPGA and performs circuit evaluation to obtain the circuit's response ($DAG(T_i;~C_n)$) to the test input $T_i$. 
We make this design decision to develop the hardware accelerator for \sys{} since acquiring the circuit’s response from a configured FPGA (circuit emulation) is significantly faster than the same process running on a host CPU (software simulation). In addition, \sys{} parallelizes the computation of circuit emulation and pipelines each step of RL process. \sys{} performs reward computation of the candidate test inputs and adaptive sampling in an online fashion to minimize data communication between the off-chip memory and the FPGA.

\vspace{-0.5em}

Note that we do not include a random number generator (RNG) in our architecture design. Instead, \sys{} stores a set of random numbers pre-computed on CPU using the inherent variation of the operating system. This design choice has two benefits: (i) The hardware overhead of a True RNG is non-trivial and not desired; (ii) Random numbers generated from the CPU typically features stronger randomness compared to the one generated on FPGA. 
The results of circuit emulation are used for computing the reward values of test inputs using Eqn.~(\ref{eq:reward}) during reward evaluation.
{Rare node evaluation and DAG distance computation in reward evaluation are parallelized by accommodating multiple Computing Engine (CE) in \sys{}’s design.} We also evenly partition the workload of each CE evenly offline.

After accumulating the reward for each candidate test input, our \textit{adaptive sampling} selects the ones with the highest rewards. This selection process is equivalent to \textit{sorting}. 
Therefore, \sys{} includes a sorting engine that permutes the key index based on their corresponding rewards. We implement a lightweight sorting engine based on the ‘even-odd sort’ algorithm~\cite{chen1978simplified} for adaptive sampling, incurring a linear runtime
overhead with the candidate test set size $M$.

Its is worth noticing that \sys{} does not deploy a central control unit to coordinate the computation flow. Instead, each design component in Figure~\ref{fig:arch} (a) follows a \textit{trigger-based control} mechanism~\cite{parashar2013triggered}. Particularly, each module is controlled by the status flag from its previous computation stage. For example, the adaptive sampling module (i.e., the sorting engine) in \sys{} begins to operate when the accumulation of the reward value is detected as completed. Our trigger-based control flow simplifies the control logic while satisfying the data dependency between different components in Figure~\ref{fig:global}. We detail the design of \sys{}'s circuit emulation and auxiliary circuitry as follows.

\vspace{-0.4em}
\subsection{\sys{} Circuit Emulation} \label{sec:hw_emulation}
\vspace{-0.6em}
We empirically observe from \sys{}'s software implementation that circuit evaluation (i.e., obtaining $DAG(T_i;C_n)$) dominates the execution time. Motivated to address the high latency issue of evaluating a circuit netlist on CPU, we propose to use \textit{circuit emulation} to improve \sys{}'s efficiency.
The first step of circuit emulation is to rewrite the netlist of the circuit under test ($C_n$) such that the values of internal nodes can be recorded by registers. The rewritten circuit is then connected with the auxiliary circuitry and mapped onto FPGA. 
In this way, we can emulate the response of the  target circuit $C_n$ for any test input by directly applying it on the circuit and collecting the corresponding values in the registers. The collected signal values are used to compute the three reward terms in Equation~(\ref{eq:reward}). 

Furthermore, \sys{} optimizes the latency of hardware evaluation by storing the emulation results in a ping-pong buffer {(i.e., consisting of two buffers denoted with $A$ and $B$)} and decoupling it from other hardware components as shown in Figure~\ref{fig:arch} (a). More specifically, the reward computing engine (CE) calculates the reward of the candidate test input using the data from 
{buffer A}. 
In the meantime, the emulator acquires the states of $C_n$ given the next input $T_i$ and stores the results into {buffer $B$.}

\vspace{-0.3em}
\subsection{\sys{} Reward Computing Engine}
\begin{figure*}[ht!]
    \centering
    \includegraphics[width=1.0\textwidth]{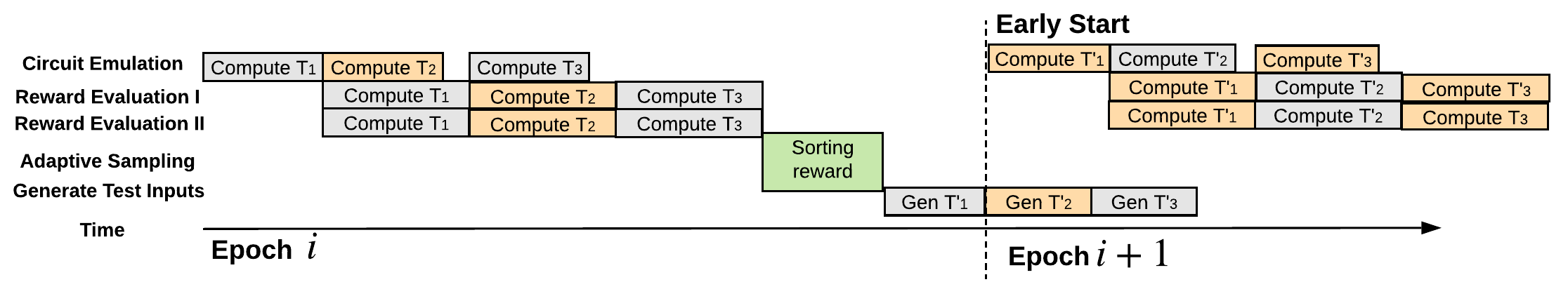}     
    \vspace{-1.2em}
    \caption{\sys{}'s hardware accelerator employs pipelining optimization to generate test patterns for HT detection.}
    \label{fig:pipeline}
\end{figure*}

\vspace{-0.5em}
\noindent \textbf{Pipeline with Early Starting.} Our architecture design aims to maximize the overlapping time between each execution stage of \sys{} to increase the throughput of TPG. 
As shown in Figure~\ref{fig:pipeline}, the ping-pong buffer enables pipelined execution of hardware emulation and reward evaluation. 
Furthermore, reward evaluation and adaptive sampling can be pipelined across different iterations. 
We can see from Figure~\ref{fig:pipeline} that epoch $(i+1)$ can start circuit emulation and reward evaluation when the previous epoch begins to generate new test inputs for the next epoch. As such, the latency of candidate test inputs generation can be hidden by circuit emulation and reward evaluation.

\noindent \textbf{Scalable Reward Computing Engine.} Once circuit emulation finishes for the current input $T_i$, \sys{} begins to calculate the reward of this test input using Equation~(\ref{eq:reward}).
From the hardware perspective, the reward term $V_{rare}$ and $V_{scoap}$ is computed by accumulating the number of activated rare nodes and the corresponding SCOAP values from the circuit $C_n$, and the reward $V_{DAG}$ is computed by accumulating the Hamming Distance (i.e., XOR) between the values in the current DAG ($DAG(T_i; C_n)$) and the historical ones ($DAG(S_i; C_n)$). 
Independence between different groups of wire signals typically exists in circuits. 
\sys{} leverages this property by distributing the computation involving independent groups of nodes to different reward computing engines as shown in Figure~\ref{fig:arch} (b).  
As such, each CE stores a subset of DAG nodes' values in the associated DAG buffer. The accumulation of the ultimate reward score completes when the last CE finishes reward computing.

\vspace{-0.6em}
\section{Evaluations}  \label{sec:eval}
\vspace{-0.8em}
We investigate \sys{}'s performance for Hardware Trojan detection on various benchmarks, including ISCAS'85~\cite{hansen1999unveiling}, MCNC~\cite{mcnc}, {and ISCAS'89~\cite{iscas89}}. The statistics of the evaluated benchmarks are summarized in Table~\ref{tab:benchmark}.
{To apply \sys{} on sequential circuits in the ISCAS'89 benchmark, we unroll the circuit for two time frames and convert it to a combinational one~\cite{arora2004enhancing,yuan2015sequential}. Note that the unrolling process duplicates the combinational logic blocks, thus increasing the effective circuit size for Trojan detection.}
The transition probability ($P_{trans}$) threshold for rare nodes is set to $P_T=0.1$ {for ISCAS'85 and MCNC benchmarks. As for two ISCAS'89 circuits, we use $P_{trans}=0.0005$ such that the number of rare nodes is at the same level as the previous two benchmarks.} The identification results are shown in the last column of Table~\ref{tab:benchmark}. 
{To compare the performance of \sys{} and other logic testing-based Trojan detection methods, we use trigger coverage and Trojan coverage as the metrics to quantify detection effectiveness. To characterize detection efficiency, we use the number of test vectors and the detection runtime as the metrics. We empirically show that \sys{} achieves a higher Trojan detection rate with shorter runtime overhead compared to the counterparts in the rest of this section.}

\vspace{-1em}
\begin{table}[ht!]
\centering
\caption{Summary of the evaluated circuit benchmarks. }
\vspace{0.4em}
\label{tab:benchmark}
\scalebox{0.98}{
\begin{tabular}{ccccccc}
\hline
Circuit& dataset & \#in & \#out & \#gate & \thead{\# of rare nodes \\$(P_{trans} < P_T)$}\\ \hline
c432& ISCAS-85 &36 &7 &160 & 14\\ \hline
c499& ISCAS-85 &41 &32 &202 & 48\\ \hline
c880& ISCAS-85  &60 &26 &383 & 74\\ \hline
c3540& ISCAS-85 &50 &22 &1669 & 218\\ \hline
c5315& ISCAS-85 &178 &123 &2307 & 169\\ \hline
c6288& ISCAS-85 &32 &32 &2416 & 245\\ \hline
c7552& ISCAS-85 &207 &108 &3512 & 266\\ \hline
des& MCNC &256 &245 &6473 & 2316\\ \hline
ex5& MCNC &8 &63 &1055 & 432\\ \hline
i9& MCNC &88 &63 &1035 & 85\\ \hline
seq& MCNC &41 &35 &3519 & 1356\\ \hline
{s5378} & {ISCAS-89} &{35} &{49} &{2958} & {258} \\ \hline
{s9234} & {ISCAS-89} &{19} &{22} &{5825} & {398} \\ \hline
\end{tabular}
}
\end{table}

\noindent \textbf{Experimental Setup.}
Adhering to our threat model defined in Section~\ref{sec:threat}, we first design the HT and insert it to each benchmark listed in Table~\ref{tab:benchmark}. 
We use a logic-AND gate as the Trojan trigger and select three rare nodes with rare value 1 as the inputs.
To fully characterize the performance of \sys{}, we devise various HTs for each circuit (i.e., using different combinations of rare nodes as the trigger) and repeat the insertion for $50$ times. 
{Our Trojaned benchmarks include `hard-to-trigger' HTs with activation probabilities around $10^{-7}$ (e.g., $c3540$). } 
To compare the performance of \sys{} with prior works, we re-implement MERO~\cite{chakraborty2009mero} and TRIAGE~\cite{nourian2018hardware} based on the methodology described in the paper using Python.
Our experiments are performed on an Intel Xeon E5-2650 v4 processor with 14.5 GiB of RAM.

\vspace{0.3em}
{\tikz\draw[black,fill=black] (-0.5em,-0.5em) rectangle (-0.2em,-0.2em);} \textbf{MERO Configuration.} 
We use the parameter selection strategy suggested in MERO~\cite{chakraborty2009mero} for re-implementation. 
Particularly, we set the size of random patterns to 2,500.
The hyper-parameter of MERO is $N$ (desired number of times that each rare node shall be activated). A large value of $N$ achieves a higher detection rate while resulting in a larger test set~\cite{chakraborty2009mero}. 
{We use $N=1,000$ in the experiments since this is the value suggested by MERO~\cite{chakraborty2009mero}.}

\vspace{0.3em}
{\tikz\draw[black,fill=black] (-0.5em,-0.5em) rectangle (-0.2em,-0.2em);} \textbf{TRIAGE Configuration.}
We use a population size of 100 and select 20 test inputs with the highest fitness score in each generation. 
The probability of crossover and mutation is set to 0.9 and 0.05, respectively. 
The termination condition in TRIAGE~\cite{nourian2018hardware} is used to evolve the test patterns.

\vspace{0.3em}
{\tikz\draw[black,fill=black] (-0.5em,-0.5em) rectangle (-0.2em,-0.2em);} \textbf{\sys{} Configuration.} 
In \sys{}'s circuit profiling stage, we use the Testability Measurement Tool~\cite{scoaptool} to compute the SCOAP parameters. 
The SAT-based smart initialization step of \sys{}'s Phase 2 is performed using the pycosat library~\cite{pycosat}.  
Our framework is developed in Python language and does not require extensive hyper-parameter tuning.
To ensure the three reward terms in Equation~(\ref{eq:reward}) have comparable values within the range of $[0,10]$, we set the hyper-parameters to $\lambda_1=0.05$, $\lambda_2=0.0001$, $\lambda_3=0.00025$.
The candidate test size and the step size in Algorithm~\ref{alg:adaptive_rl} are set to $M=200$ and $L=80$ for all benchmarks, respectively. 
{We use the percentage threshold $p=95\%$ to identify rare nodes and set the target activation times to $N=20$.}
{The maximal iteration time is set to $I_{max}=500$.}  

\begin{figure}[b!]
    \centering
    \includegraphics[width=0.44\textwidth]{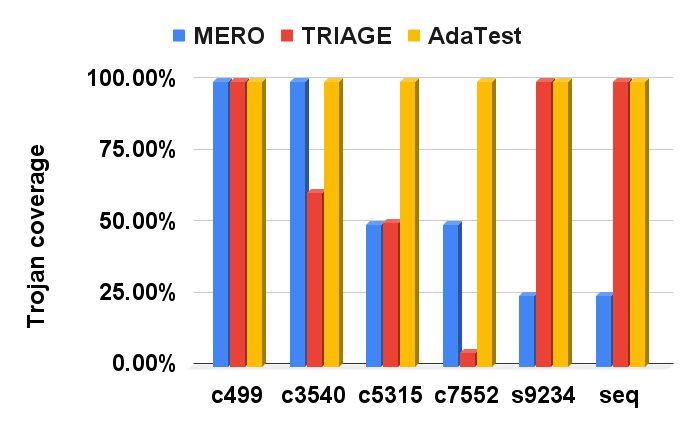} 
    \vspace{-0.8em}
    \caption{Trojan detection rates of \sys{} and prior works on various benchmarks. }
    \label{fig:effect_comp}
\end{figure}

According to the performance metrics in Section~\ref{sec:global}, we use the \textit{trigger coverage} (percentage of trigger nodes identified by the test set) and the \textit{Trojan coverage} (i.e., detection rate) to quantify the effectiveness of HT detection.
Meanwhile, we measure the \textit{test set generation time} and test set size of each technique for efficiency comparison. 
To obtain an accurate and comprehensive performance measurement, we design $50$ different HTs for each benchmark in Table~\ref{tab:benchmark} while fixing the number of trigger nodes to {$3$}. 
Each set of devised HTs is inserted into the circuit independently. 
{We run \sys{} detection on each Trojaned circuit for $20$ times. 
The trigger and Trojan coverage for each benchmark are computed as the average value over $50 \times 20=1000$ runs. }

\begin{table}[b!]
\centering
\caption{Performance comparison summary of different Trojan detection techniques. }
\vspace{0.2em}
\label{tab:detect_perf}
\scalebox{0.85}{
\begin{tabular}{|c|l|r|r|r|r|}
\hline
circuit & Method  & \multicolumn{1}{l|}{\# test vectors} & \multicolumn{1}{l|}{Runtime (s)} & \multicolumn{1}{l|}{Trigger coverage} & \multicolumn{1}{l|}{Trojan coverage} \\ \hline
\multirow{3}{*}{c499}  & {MERO}    & 1660      & 136.49   & 100.00\%  & 100.00\%   \\ \cline{2-6} 
 & TRIAGE  & 250000    & 25.91  & 100.00\%   & 100.00\%   \\ \cline{2-6}  & AdaTest & \textbf{1010}   & 13.60   & 100.00\%  & 100.00\%  \\ \hline
\multirow{3}{*}{c880}  & {MERO}    & 1332   & 352.54   & 100.00\%  & 100.00\%                     \\ \cline{2-6}  & TRIAGE  & 250000 & 1.75    & 82.29\% & 18.00\%   
\\ \cline{2-6} & AdaTest & \textbf{429}    & 0.43     & 100.00\%      & 97.50\%                     \\ \hline
\multirow{3}{*}{c3540} & {MERO}    & 1920   & 1577.36        & 100.00\%           & 100.00\%    \\ \cline{2-6}  & TRIAGE  & 250000     & 25.85       & 100.00\%  & 61.00\%                          \\ \cline{2-6} & AdaTest & \textbf{905}  & 22.61        & 100.00\%  & \textbf{100.00\%}    \\ \hline
\multirow{3}{*}{c5315} & {MERO}    & 9265   & 1660   & 100.00\%    & 50.00\%     
\\ \cline{2-6}  & TRIAGE  & 250000  & 37.14    & 100.00\%      & 50.50\%                              \\ \cline{2-6}  & AdaTest & \textbf{1300}     & 19.76   & 100.00\%     & \textbf{100.00\%} \\ \hline
\multirow{3}{*}{c6288} & {MERO}    & 1906  & 1867.57 & 100.00\%    & 100.00\%                   \\ \cline{2-6}  & TRIAGE  & 250000   & 44.11     & 100.00\%    & 91.50\%                              \\ \cline{2-6}  & AdaTest & \textbf{900}    & 47.06 & 100.00\%   & \textbf{99.50\%}      \\ \hline
\multirow{3}{*}{c7552} & {MERO}    & 1916     & 18650.5             & 100.00\%                               & 50.00\%         \\ \cline{2-6}  & TRIAGE  & 250000   & 20.93                                  & 93.88\%                               & 5.00\%                               \\ \cline{2-6}  & AdaTest & \textbf{1600}   & 39.79   & 98.08\%   & \textbf{100.00\%}    \\ \hline
\multirow{3}{*}{{s5378}} & {MERO}    & 1103     & 30960.11        & 100.00\%                               & {100.00}\%                              \\ \cline{2-6}  & TRIAGE  & 300           & 0.45                  & 100.00\%         & 100.00\%          \\ \cline{2-6}   & AdaTest & {100}   & 11.58     & 100.00\%      & \textbf{100.00\%}      \\ \hline
\multirow{3}{*}{{s9234}} & {MERO}   & 11    & 29737.84     & 100.00\%        & 25.00\%    \\ \cline{2-6}  & TRIAGE  & 500                                  & 35.625
                      & 100.00\%                               & 100.00\%                               \\ \cline{2-6} 
                      & AdaTest & 140                                    & 124.99                                 & 100.00\%                               & \textbf{100.00\%}                    \\ \hline
\multirow{3}{*}{des}   & {MERO}    & 1120                                  & 34943.41                                  & 100.00\%                               & 100.00\%                           \\ \cline{2-6}   & TRIAGE  & 2500    & 0.84                                   & 100.00\%             & 100.00\%      
\\ \cline{2-6} & AdaTest & \textbf{156.8}             & 15.11  & 92.88\% & \textbf{100.00\%}   \\ \hline
\multirow{3}{*}{ex5}   & {MERO}    & 904     & 115.22    & 100.00\%       & 100.00\%      \\ \cline{2-6}  & TRIAGE  & 2500  & 0.13     & 99.13\%         & 100.00\%    \\ \cline{2-6} & AdaTest & \textbf{500}     & 12.35      & 93.81\%    & \textbf{100.00\%}   \\ \hline
\multirow{3}{*}{i9}    & {MERO}    & 268                                  & 808.56                                  & 100.00\%                               & 100.00\%                             \\ \cline{2-6}  & TRIAGE  & 2500    & 0.09     & 100.00\%          & 100.00\%     \\ \cline{2-6}  & AdaTest & 600     & 12.15                                  & 94.58\%                               & \textbf{100.00\%}                    \\ \hline
\multirow{3}{*}{seq}   & {MERO}    & 1776   & 3773.3    & 100.00\%                               & 66.67\%          \\ \cline{2-6} & TRIAGE  & 250000      & 22.11      & 95.44\%                               & 2.00\%                               \\ \cline{2-6}  & AdaTest & 3700    & 20.72     & 94.58\%     & \textbf{82.00\%}    \\ \hline
\end{tabular}
}
\end{table}

\vspace{-0.5em} 
\subsection{Detection Effectiveness}  \label{sec:effect}
\vspace{-0.5em}
We assess the detection performance of \sys{}, MERO, and TRIAGE using the aforementioned experimental setup. 
Figure~\ref{fig:effect_comp} compares the Trojan coverage of the three HT detection techniques on different benchmarks.
One can see that our framework achieves uniformly higher detection rates across various circuits. 
The superior HT detection performance of \sys{} is derived from our definition of \textit{adaptive}, \textit{context-aware} reward functions in Equation~(\ref{eq:reward}).

We use two metrics to quantitatively compare the effectiveness of different HT detection techniques: trigger coverage rate and Trojan detection rate.
{Note that \sys{} determine a Hardware Trojan is present in the circuit if the set of test patterns generated using Alg.~\ref{alg:adaptive_rl} result in Trojan activation when the test inputs are applied on the circuit. Therefore, our detection method does not have any false positives and we focus on evaluating the detection rates (which corresponds to the false negative rate).}
Table~\ref{tab:detect_perf} summarizes the HT detection results of three different methods on the benchmarks in Table~\ref{tab:benchmark}. 
The trigger coverage and Trojan coverage results are shown in the last two columns of Table~\ref{tab:detect_perf}.
It can be seen that \sys{} achieves the highest Trojan coverage while requiring the shortest test generation time across most of the benchmarks.
More specifically, \sys{} achieves an average of 
{$15.61\%$ and $29.25\%$}
Trojan coverage improvement over MERO~\cite{chakraborty2009mero} and TRIAGE~\cite{nourian2018hardware}, respectively. 
The superior HT detection performance of our logic testing-based approach is derived from the diverse test patterns found by \sys{} adaptive RL-driven input space exploration technique (see Section~\ref{sec:adaptive_rl}).
We not only encourage the activation of rare nodes and differentiate their qualities using SCOAP testability parameters, but also explicitly characterize the graph-level distance of the CUT status under different test stimuli. 

\begin{figure}[ht!]
    \centering
    \includegraphics[width=0.44\textwidth]{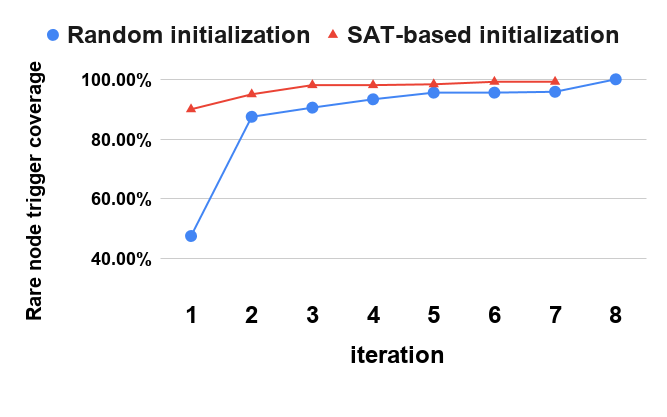} 
    \vspace{-2em}
    \caption{The rare node coverage of \sys{} versus the number of executed iterations on c3540 benchmark. }
    \label{fig:initialization}
\end{figure}

We measure the dynamic rare node coverage 
versus the number of executed iterations to validate the \textit{time-evolving} property of \sys{} framework. 
Figure~\ref{fig:initialization} shows the coverage results of \sys{} with random initialization and SAT-based smart initialization on the $c3540$ benchmark. 
We can make two observations from Figure~\ref{fig:initialization}: (i) \sys{} consistently improves the rare node coverage over time (with either initialization method); 
(ii) SAT-based smart initialization improves the convergence speed of \sys{}, thus reducing our test set generation time. 
The first observation corroborates the efficacy of our RL-based \textit{progressive} test pattern generation method. 
The second observation reveals the importance of proper initialization for fast convergence of RL exploration.
Note that a shorter convergence time (i.e., a smaller number of iterations in Algorithm~\ref{alg:adaptive_rl}) indicates s smaller test set returned by \sys{}, which is beneficial to reduce the test generation time for higher detection efficiency. 

\vspace{-1em}

\vspace{-0.7em}
\subsection{Detection Efficiency} \label{sec:effic}
\vspace{-0.3em}

We characterize the efficiency of \sys{} for logic testing based-HT detection using two metrics: the test set size (\textit{space efficiency}), and the test set generation time (\textit{runtime efficiency}). 
The quantitative efficiency measurements of three HT detection methods are shown in the third and fourth columns of Table~\ref{tab:detect_perf}. 
It can be computed that \sys{} engenders an average of 
{2.04$\times$} and {155.04$\times$}  
reduction of test set size compared to MERO and TRIAGE across all benchmarks, respectively.  
The reduction of test set size has two benefits: (i) A smaller test set features a lower memory footprint; (ii) For on-chip test pattern generation, a smaller test set suggests a shorter test generation time.

\begin{figure}[ht!]
    \centering
    \includegraphics[width=0.45\textwidth]{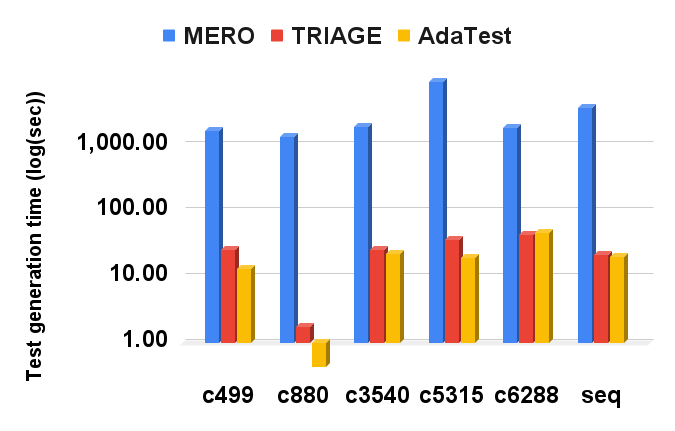} 
    \vspace{-1.2em}
    \caption{Test set generation time comparison between \sys{} and prior works. The runtime shown by the y-axis is represented in the log scale. }
    \label{fig:effic_comp}
\end{figure}

Figure~\ref{fig:effic_comp} compares the required test generation time of \sys{}, MERO, and TRIAGE to achieve the coverage results on various benchmarks in Table~\ref{tab:detect_perf}.
Note that we use \textit{log-scale} for the vertical axis since the range of runtime is diverse across different circuits. 
We can observe that \sys{} is the most efficient HT detection method among the three and it also achieves high Trojan coverage (last column of Table~\ref{tab:detect_perf}). 
{More specifically, \sys{} engenders an average of 
{366.26$\times$} and {0.63$\times$} test generation speedup compared to MERO~\cite{chakraborty2009mero} and TRIAGE~\cite{nourian2018hardware}, respectively.} {Note that although the runtime of TRIAGE is smaller, its Trojan detection rate is ~$30\%$ lower than \sys{}.}


\vspace{-1.1em}
\vspace{0.5em}

\vspace{-0.8em}
\subsection{\sys{} Architecture Evaluation} 
\begin{figure}[ht!]
    \centering
    \includegraphics[width=0.57\textwidth]{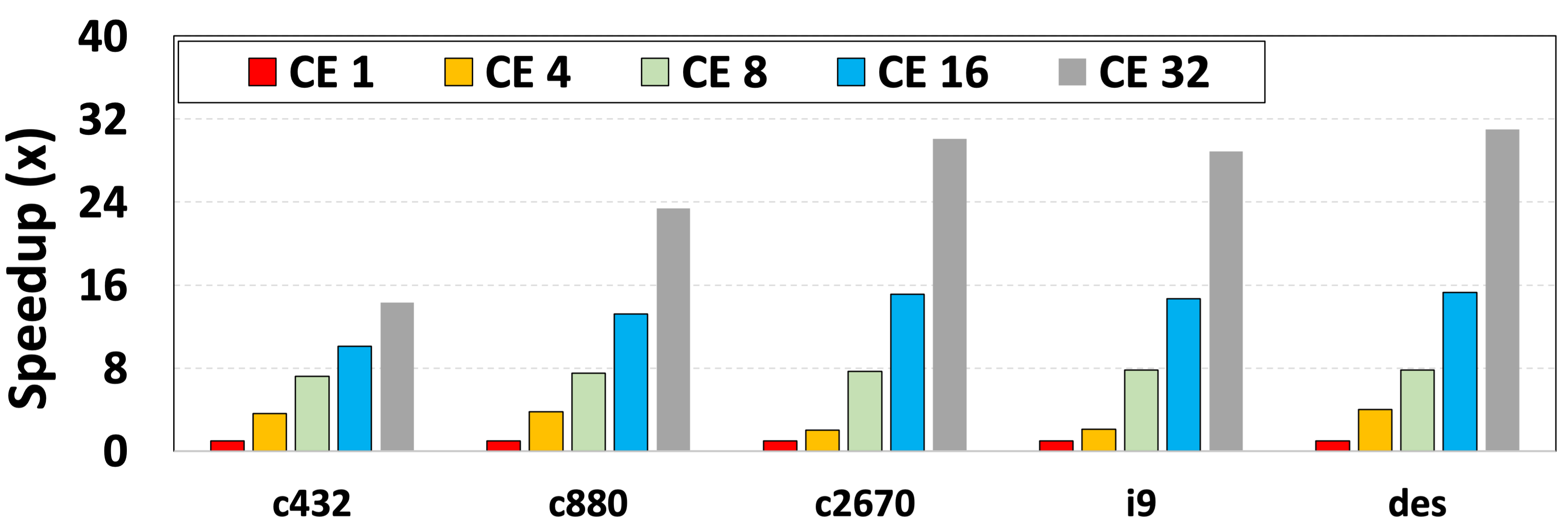} 
    \vspace{-0.3em}
    \caption{\sys{}'s scalability to the number of DAG reward computing engines. The speedup is near-linear with $N_{CE}$ on large circuits where reward evaluation is the computation bottleneck.}
    \label{fig:DAG_CE}
\end{figure}

\vspace{-0.5em}
The resource utilization of \sys{} depends on the input length and the circuit size. We report the resource utilization results of the evaluated benchmarks in Table~\ref{tab:workload}.  
Figure~\ref{fig:DAG_CE} shows that \sys{} architecture achieves approximately linear speedup w.r.t. to the number of CEs. Our hardware design can be scaled up by adding more reward computing engines to parallel the circuit emulation process as \sys{}'s computation bottleneck is reward evaluation of the test patterns. Nevertheless, the speedup saturates when $N_{CE}$ is sufficiently high. 
\sys{} broadcasts the wire values of the circuit response (given a test input) to all CEs via a shared data bus. Each CE scans the DAG buffer and obtains the broadcast wire values to compute the corresponding reward.  
Therefore, increasing the number of CEs does not lead to extra wire delay. However, more CEs suggests a higher overhead during reward accumulation.

\vspace{-1em}
\begin{table}[ht!]
  \caption{Resource utilization of the auxiliary circuitry on \textit{c432},\textit{c880}, \textit{c2670} and \textit{des} benchmarks with default settings ($N_{CE}=16$) on Zynq ZC706.}
  \label{tab:workload}
  \vspace{0.5em}
  \centering
  \scalebox{0.85}{
  \begin{tabular}{ccccc}
    \toprule
    Benchmarks & c432  & c880 & c2670 & des\\
    \midrule 
    BRAMS & 26 & 36 & 65 & 237 \\
    DSP48E1 & 0& 0 & 0 & 0\\
    \thead{KLUTs  (emulator usage)} & 14.9 (0.5) & 25.5 (0.6) & 61.1 (3.5) & 267.9 (26.1)\\
    \thead{FFs  (emulator usage)} & 4,440 (80) & 5,743 (160) &6,717 (317) & 12,943 (1190)\\
  \bottomrule
\end{tabular}
}
\end{table}

\vspace{-1em}




\section{Conclusion}  \label{sec:conclus}
\vspace{-0.7em}

In this paper, we present a holistic solution to Hardware Trojan detection using adaptive, reinforcement learning-based test pattern generation.
To formulate logic testing-based HT detection as an RL problem, we design an innovative reward function to characterize the quality of a test pattern from both static and dynamic aspects.  
\sys{} progressively expands the test set by identifying test input vectors with high reward values in an iterative approach.
\sys{} integrates adaptive sampling to identify and encourage high-reward test patterns, thus accelerating our RL-based input space exploration. 
We devise \sys{} using a Software/Hardware co-design approach. Particularly, we develop a domain-specific systems-on-chip architecture for efficient hardware implementation of \sys{}. Our architecture optimizes reward evaluation via circuit emulation and pipelines the computation of \sys{}. 
We perform extensive evaluations of \sys{} on various benchmarks and compare its performance with two counterparts, MERO and TRIAGE. 
Empirical results corroborate that \sys{} achieves superior effectiveness, efficiency, and scalability for HT detection compared to prior works. 
\sys{} is a \textit{generic} test pattern generation framework, we plan to investigate its performance on other hardware security problems such as logic verification and built-in self-test in our future work. 




\bibliographystyle{unsrt}
\bibliography{refs}
\end{document}